\documentclass[10pt,twocolumn,letterpaper]{article}

\usepackage{cvpr}
\usepackage{graphicx}
\usepackage{comment}
\usepackage{amsmath,amssymb} % define this before the line numbering.
\usepackage{color}
\usepackage{times}
\usepackage{epsfig}
\usepackage{diagbox}
\usepackage{afterpage}
\usepackage{makecell}
\usepackage{lineno}
\usepackage{amsfonts}       % blackboard math symbols
\usepackage{amsthm}   %\newtheorem{proof}{Proof}

\usepackage{bm}
\usepackage{algorithm}
\usepackage{algorithmic}
\newtheorem{assumption}{Assumption}
\newtheorem{theorem}{Theorem}
\usepackage{amssymb}
\newtheorem{lem}{Lemma}

\newcommand{\be}{\begin{eqnarray}}
\newcommand{\ee}{\end{eqnarray}}
\newcommand{\beq}{\begin{equation}\begin{aligned}}
\newcommand{\eeq}{\end{aligned}\end{equation}}
\newcommand{\beqn}{\begin{equation*}\begin{aligned}}
\newcommand{\eeqn}{\end{aligned}\end{equation*}}
\newcommand{\ben}{\begin{eqnarray*}}
\newcommand{\een}{\end{eqnarray*}}
\newcommand{\bena}{\begin{eqnarray*}\begin{aligned}}
\newcommand{\eena}{\end{aligned}\end{eqnarray*}}
\newcommand{\bea}{\begin{eqnarray}\begin{aligned}}
\newcommand{\eea}{\end{aligned}\end{eqnarray}}
\usepackage{algorithm}
\usepackage{algorithmic}
\usepackage{multirow}
\usepackage{booktabs}       % professional-quality tables

\usepackage[pagebackref=true,breaklinks=true,letterpaper=true,colorlinks,bookmarks=false]{hyperref}

\usepackage{authblk}
\cvprfinalcopy
\ifcvprfinal\fi
\begin{document}

\title{Towards GANs' Approximation Ability} % Replace with your title
\author{Xuejiao Liu}
\author{Yao Xu}
\author{Xueshuang Xiang\thanks{Corresponding author: xiangxueshuang@qxslab.cn}}
\affil{\normalsize Qian Xuesen Laboratory of Space Technology\\ China Academy of Space Technology}
\date{}

%\twocolumn[
%\begin{@twocolumnfalse}
\maketitle
\begin{abstract}
	Generative adversarial networks (GANs) have attracted intense interest in the field of generative models. However, few investigations focusing either on the theoretical analysis or on algorithm design for the approximation ability of the generator of GANs have been reported. This paper will first theoretically analyze GANs' approximation property. Similar to the universal approximation property of the fully connected neural networks with one hidden layer, we prove that the generator with the input latent variable in GANs can universally approximate the potential data distribution given the increasing hidden neurons. Furthermore, we propose an approach named stochastic data generation (SDG) to enhance GANs' approximation ability. Our approach is based on the simple idea of imposing randomness through data generation in GANs by a prior distribution on the conditional probability between the layers. SDG approach can be easily implemented by using the reparameterization trick. The experimental results on synthetic dataset verify the improved approximation ability obtained by this SDG approach. In the practical dataset, four GANs using SDG can also outperform the corresponding traditional GANs when the model architectures are smaller. 
	%In the practical dataset, the NSGAN/WGANGP with SDG can also outperform traditional GANs with little change in the model architectures. 
\end{abstract}
%\end{@twocolumnfalse}
%]

\section{Introduction}
Since they were first proposed by \cite{goodfellow2014generative}, there has been an explosive growth in the studies on the well-known generative adversarial networks (GANs) \cite{Salimans2016Improved,arjovsky2017wasserstein,Arora2017Generalization,Karras2018Progressive}. GANs are a new framework for estimating generative models via an adversarial process. By simulating the adversarial process between the generative model and the discriminative model, GANs can learn deep representations without extensively annotated the training data and learn the style of a group of images. Samples from the latent space are randomly selected as inputs of the generative model to learn the uncertainty of the target dataset.

Despite the great empirical success for the use of GANs in many  practical applications, many theoretical issues related to GANs are still unsolved. Most theoretical works on GANs have focused either on the design of the objective function \cite{Nowozin2016fGAN,salimans2018improving,Mao2016,Lim2017,Bellemare2017,Kodali2017,Mescheder2018Which,Miyato2018} or on the convergence of the adversarial process \cite{Liu2017Approximation,Arora2017Generalization,Arora2018Do,Bai2019Approximability}. For instance, Wasserstein GANs (WGAN) \cite{arjovsky2017wasserstein} focus on the measurements of the distance or divergence between the  real distributions and model distributions and define an approximation of the Earth mover distance but, in some cases, may still generate only poor samples or fail to converge. Then, an alternative to clip weights was proposed, where the norm of the gradient of the critic was penalized with respect to its input (WGANGP) \cite{Gulrajani2017Improved}. The works of \cite{MeschederNG17a,NagarajanK17Gradient,Mescheder2018Which} focus on the local convergence of GANs training for absolutely continuous data. \cite{Mescheder2018Which} further show that for the more realistic case of the distributions that are not absolutely continuous, and unregularized GANs training is not always convergent. Furthermore, their analysis shows that GANs training with instance noise or zero-centered gradient penalties converges and WGAN/WGANGP with a finite number of discriminator updates per generator update do not always converge to the equilibrium point.  \cite{Liu2017Approximation} focus on the theoretical issues related to GANs' approximation and convergence. The work of \cite{Liu2017Approximation} showed that if the objective function is an adversarial divergence function with some additional conditions, then the use of a restricted discriminator family has a moment-matching effect. The theoretical work of \cite{Arora2017Generalization} considered the generalization and equilibrium in GANs and suggested a dilemma for the statistical properties of GANs: powerful discriminators lead to overfitting, while weak discriminators cannot detect mode collapse.

To the best of our knowledge, few investigations have focused on the approximation ability of the GANs generator. A basic question of the approximation ability is as follows: \textbf{can the generator approximate the target data distribution?} Once the approximation analysis of the generator and the convergence analysis of the adversarial learning process are carried out, we can estimate the error between the true data distribution and the learned generator distribution. It is commonly thought that an increase in the model size of the generator allows for the enhancement in its representation ability, yielding an increasing approximation ability \cite{goodfellow2014generative}. This paper will first theoretically answer the above basic question. Similar to the universal approximation property (UAP) of the fully connected neural networks with a single hidden layer, we can prove that the generator with input latent variable in GANs can universally approximate the potential data distribution given the increasing hidden neurons.

Another approach for the investigation of the generator's approximation ability is to design effective model architectures or training strategy \cite{radford2016unsupervised,Dumoulin2017,Zhang2018,Karras2018A,Karras2019,Brock2019Large}. For instance, \cite{radford2016unsupervised} designed the deep convolutional GAN that shows great performance in unsupervised learning. \cite{Bengio2013Deep,Karras2018A,Brock2019Large,Chen2019On} designed a new type of generator that introduces noise to the hidden layers to enhance the representation ability of the generators. \cite{Karras2018Progressive} proposed a new training methodology for the progressively growth of both the generator and discriminator: starting from a low resolution, new layers that model increasingly fine details as the training progresses are added. The present paper will propose an approach, named stochastic data generation (SDG), that introduces randomness during the feedforward process of the generator to enhance GANs' approximation ability. The underlying idea is simply to impose a prior distribution on the conditional probability between the layers. Using the reparameterization trick, SDG is easily implemented. The experimental results on a synthetic dataset verify the improved  approximation ability of SDG. In addition, the experimental results on the practical datasets (MNIST, CIFAR10 and CELEBA) show that the NSGAN \cite{goodfellow2014generative}, DRAGAN \cite{Kodali2017}, WGANGP \cite{Gulrajani2017Improved} or LSGAN \cite{Mao2016} with SDG also outperforms the corresponding traditional ones when the model architectures are smaller. 

\section{UAP of GANs' Generator}
We make a basic assumption for the target data distribution. 
\begin{assumption}\label{assump1}
	For the target data distribution $p_{\bf X}({\bf x})$, ${\bf x} \in K \subseteq \mathbb{R}^D$, we assume that:
	\begin{enumerate}
		\item $K$ is a compact set;
		\item $p_{\bf X}({\bf x})$ is continuous on $K$;
		%\item $d:={\rm dim}(K) \leq D$.
		\item There exists random variables $x^1$, $x^2$, $\cdots$, $x^d$ $\in \mathbb{R}$ are independent of each other, $d \leq D$ and $p_{\bf X}({\bf x}) = p_{\bf X}(x^1,x^2,\cdots,x^d)$.
		%There exists random variables $\bf x^1$, $\bf x^2$, $\cdots$, $\bf x^d$ $\in \mathbb{R}^D$ are independent of each other s.t. ${\bf x}$ $\in$ $span(\bf x^1, x^2, \cdots, x^d)$, $\forall\, {\bf x} \in K$. That is $p_{\bf X}({\bf x}) = p_{\bf X}(\bf x^1,\bf x^2,\cdots,\bf x^d)$.
	\end{enumerate}
\end{assumption}

Define the generator as $G({\bf z};\theta)$ with ${\bf z} \in \mathbb{R}^d$, and we assume ${\bf z} \sim U^d(0,1)$, with the uniform distribution on $(0,1)^d \subset \mathbb{R}^d$. This section will prove that $p_{G}$ can approximate an arbitrary data distribution $p_{\bf X}$, i.e., the universal approximation property of the generator of GANs. 

Suppose that the architecture of generator $G$ is a fully connected neural network with one hidden layer, redefined by $G({\bf z};\theta,N):= \mathbf{C} \sigma ( \mathbf{W} {\bf z} + \mathbf{b})$, where $N$ is the number of neurons in the hidden layer, $\theta:=\{\mathbf{C}, \mathbf{W}, \mathbf{b}\}$, $\mathbf{C} \in \mathbb{R}^{D \times N}, \mathbf{W} \in \mathbb{R}^{N \times d}, \mathbf{b} \in \mathbb{R}^{N}$ and $\sigma(\cdot)$ is the pointwise activation function. First, we recall the UAP of a general neural network for approximating a target function \cite{cybenko1989approximation,barron1993universal,leshno1993multilayer}. 
\begin{lem}\label{lem:1}
	For any continuous function $f({\bf x})$ on a compact set $A$, we have, as $N\rightarrow \infty$, 
	\bea
	|G({\bf x};\theta,N) - f({\bf x})| \rightarrow 0, \quad \forall\, {\bf x} \in A.
	\eea
	%There exists a feedforward neural network, having only a single hidden layer, which uniformly approximates $f$ to within an arbitrary $\epsilon>0$ on K.
\end{lem}
This paper will prove a consistent UAP of the generator in GANs as follows:
\begin{theorem}\label{thm:1}
	For any data distribution $p_{\bf X}({\bf x})$ with Assumption \ref{assump1}, we have, as $N\rightarrow \infty$, 
	\bea
	|p_{{G}({\bf z};\theta,N)}({\bf x}) - p_{\bf X}({\bf x})| \rightarrow 0, \quad \forall\, {\bf x} \in K, 
	\eea
	where $p_{{G}({\bf z};\theta,N)}({\bf x}) = \mathbb{E}_{{\bf z} \sim U^d(0,1)}\,\delta({\bf x}-{G}({\bf z};\theta,N))$. 
\end{theorem}
Here, $p_{{\bf G}({\bf z};\theta,N)}({\bf x})$ refers to the distribution obtained by a generative model with one hidden layer. It is important to note the difference between Lemma \ref{lem:1} and Theorem \ref{thm:1}: the former focuses on the approximation of the function, while the latter aims to approximate a density function. 

To prove Theorem \ref{thm:1}, we first introduce a key lemma. 

\begin{lem}\label{lem:2}
	For any data distribution $p_{X}({\bf x})$ with Assumption \ref{assump1}, 
	there exists a function $Q(\bf z)$, ${\bf z} \sim U^d(0,1)$, such that 
	\begin{enumerate}
		\item $p_{Q(\bf z)}({\bf x}) = p_{\bf X}({\bf x})$, $\forall\, {\bf x} \in K$;
		\item $Q(\bf z)$ is a continuous function;
	\end{enumerate}
	where $p_{Q(\bf z)}({\bf x}) = \mathbb{E}_{{\bf z} \sim U^d(0,1)}\,\delta ({\bf x} - Q(\bf z))$.
\end{lem}

See the Appendix for a constructive proof. 
For 1D data with $D=1$, we can relieve the assumption on the distribution of $\mathbf{z}$. 
The key contribution of Lemma~\ref{lem:2} is to bridge density function $p_{\bf X}({\bf x})$ to a function $Q(\bf z)$. 
Let us address the proof of Theorem \ref{thm:1}. 
\begin{proof}
	By Lemma \ref{lem:2}, there exists a continuous function $Q({\bf z})$, ${\bf z} \sim U^d(0,1)$, such that 
	$p_{Q({\bf z})}({\bf x}) = p_{\bf X}({\bf x})$, $\forall\, {\bf x} \in K$. 
	Apparently, $Q({\bf z})$ is defined on a compact set. 
	Thus, by Lemma \ref{lem:1}, we have that for an arbitrary $\epsilon>{\bf 0}$, 
	there exists $N \in \mathbb{Z}^{+}$ such that $|G({\bf z};\theta,N) - Q({\bf z})|\leq \epsilon$, $\forall\, {\bf z} \in (0,1)^d$. 
	Letting $\epsilon_{\bf z} = G({\bf z};\theta,N) - Q({\bf z})$, we have $|\epsilon_{\bf z}| \leq \epsilon$. 
	With the same notation as that in the proof of Lemma \ref{lem:2}, we have
	\bena
	&p_{{G}({{\bf z}};\theta,N)}({\bf x}) = \mathbb{E}_{{\bf z} \sim U^d(0,1)}\,\delta ({\bf x} - {G}({\bf z};\theta,N)) \\
	&= \int_{(0,1)^d}\delta({\bf x}-Q({\bf z})-\epsilon_{\bf z})d{\bf z} \\
	&= \int_{0}^{1}\ldots\int_{0}^{1}\delta((x^1-Q(\lambda^1),\ldots, x^d - Q(\lambda^d))-\epsilon_{\bf z})\\
	&\quad (k_2^1-k_1^1)p_{X_1}(k_1^1+\lambda^1(k_2^1-k_1^1))\cdots\\
	&\quad (k_2^d-k_1^d)p_{X_d}(k_1^d+\lambda^d(k_2^d-k_1^d))d\lambda^1\ldots d\lambda^d \\
	%&= \int_{0}^{1}\delta(x-k_1-\lambda(k_2-k_1)-\epsilon_{\bf z}) \\
	%&\quad \quad p_X(k_1+\lambda(k_2-k_1)) d\lambda \\
	&=p_{\bf X}({\bf x}-\epsilon_{\bf z}).
	\eena
	
	Then, we obtain
	\bena
	|p_{{G}({\bf z};\theta,N)}({\bf x}) - p_{\bf X}({\bf x})| &= |p_{\bf X}({\bf x}-\epsilon_{\bf z}) - p_{\bf X}({\bf x})| \\
	&\leq C \epsilon_{\bf z} \leq C \epsilon,
	\eena
	where we use Assumption \ref{lem:1}.2. 
	
	This completes the proof. 
\end{proof}

Now, we give some remarks about our theoretical results:
\begin{itemize}
	\item Actually, the assumption for the independence of the potential variables of data is somehow strict. 
	For a general data distribution when $D>1$, the proof of the UAP for generator is an open problem. 
	Technically speaking, we will consider fix this problem from following aspects: modifying the construction strategy to make it adoptable for general data distribution, directly estimating the error between $p_{\bf X}(x^1,x^2)$ and $p_{X_1}(x^1)p_{X_2}(x^2)$, or proving it from the perspective of functional analysis. 
	\item As we can easily check, the key point for proving Theorem \ref{thm:1} is Lemma \ref{lem:2}, which translates approximating a density function to approximating a function. Combining with the UAP of the neural network with one hidden layer, we can expect the UAP of the generator. 
	That means once the generator has the ability of UAP for approximating a given function, like CNN architecture or MLP, we would obtain the consistent theoretical conclusion, i.e. Theorem \ref{thm:1}. 
\end{itemize}

%\subsection{General case}

\section{Stochastic Data Generation}
It is known that the density function of a generator $p_G(\mathbf{x}) = \mathbb{E}_{\mathbf{z} \sim p_\mathbf{z}} p_G(\mathbf{x} | \mathbf{z})$. 
For the traditional GANs, the generator is expressed as a deterministic feedforward network with $p_G(\mathbf{x} | \mathbf{z}) = \delta(\mathbf{x} - G(\mathbf{z}))$. 

However, in reality, when facing the same environment in different times, a heuristic concept is that cells, animals or even human beings do not always react in an identical way; i.e., the processing of received signals from environment in organisms will not remain constant. When the neuron system receives identical signals at different times, the neurons will be activated, yet usually, the intervals between the spikes vary randomly, and the brain encodes the information differently \cite{stein2005neuronal,denfield2018attentional,goris2014partitioning}. This means that the randomness of the networks may arise from the neuron itself; i.e., the generator should not be a deterministic process. 

Actually, the stochastic neural network (SNN) is indeed a network with an intrinsic randomness~\cite{Rezende2014Stochastic,Blundell2015Weight,Gal2016Dropout}. In these networks, a prior probability distribution over the weights is introduced, and  the stochastic neural network is applied for the classification or regression problems. The key issue of SNN is to approximate the posterior distribution $p(w|X)$, where $w$ are the weights of the neural network and $X$ is the given dataset. The addition of the prior distribution over the weights is used to impose a constraint and regularization to the network, thereby mitigating the occurrence of network overfitting. In conclusion, SNN focuses on classification or regression problems and actually decreases the NN's approximation ability. 

For GANs, several approaches have been reported~\cite{Bengio2013Deep,Karras2018A,Brock2019Large,Chen2019On} that impose the noise on the hidden layers of the generative models. 
They empirically introduce some noise within the activation of hidden layers. 
For instance, self-modulation~\cite{Chen2019On} modulated the hidden layers as a function 
of input noise ${\bf z}$. However, once the input is fixed, the generator is also a deterministic feedforward process, 
i.e., there is no randomness in $p_G(\mathbf{x} | \mathbf{z})$. 
%However, the independent contribution of the added noise on the hidden layers is not clear. 
Here we propose another approach to introduce randomness in the hidden layers of the generator. 
Our key objective is to make sure there exists randomness in $p_G(\mathbf{x} | \mathbf{z})$. 
Suppose that the generative process in a traditional G is 
$${\bf z} \xrightarrow{M^0} {\bf h}^1 \xrightarrow{M^1} {\bf h}^2 \xrightarrow{M^2} \cdots \xrightarrow{M^L} \mathbf{x}, $$
where ${\bf h}^i$ is the $i$-th hidden layer with $N_i$ neurons, particularly ${\bf h}^0 = \bf z$. 
As mentioned above, the output of the $(i+1)$-th hidden layer is $p_G({\bf h}^{i+1} | {\bf h}^{i}) = \delta({\bf h}^{i+1} - M^{i}({\bf h}^{i}))$, where $M^{i}$ is the mapping from layer $i$ to layer $i+1$.
The input dimension of ${\bf z}$ is $d$, and the output dimension of ${\bf x}$ is $D$. 

Here, we impose a prior Gaussian distribution on the conditional probability between the layers; i.e., 
$$p({\bf h}^{i+1} | {\bf h}^{i}) \sim \mathcal{N}(\mu^i_{{\bf h}^{i}}, \sigma^i_{{\bf h}^{i}}), \mu^i_{{\bf h}^{i}} = M^i_\mu({\bf h}^{i}), \sigma^i_{{\bf h}^{i}} = M^i_\sigma({\bf h}^{i}),$$ 
where $\mu^i_{{\bf h}^{i}},\sigma^i_{{\bf h}^{i}}$ that have the same size as ${\bf h}^{i+1}$, $M^i_\mu,M^i_\sigma$ are the parameterized mapping, just like $M^i$. 
Then, the generative process in the stochastic data generation SDG is described by
$${\bf z} \xrightarrow{M_\mu^0,M_\sigma^0} {\bf h}^1 \xrightarrow{M_\mu^1,M_\sigma^1} {\bf h}^2 \xrightarrow{M_\mu^2,M_\sigma^2} \cdots \xrightarrow{M^L} \mathbf{x}. $$
Note that given a fixed input ${\bf z}^*$, ${\bf h}^{1} \sim  \mathcal{N}(M_\mu^0({\bf z}^*),M_\sigma^0({\bf z}^*))$. 
That means $p_G(\mathbf{x} | \mathbf{z})$ is a stochastic process, not a deterministic process~\cite{Chen2019On}. 
In the experiments, we will compare the results of the traditional G and stochastic data generation SDG, and show that the improvement in the approximation ability of SDG approach is due to the introduction of a prior distribution on the conditional probability rather than due to the increase in the number of network parameters. 

\textbf{UAP of SDG}. 
We note that the traditional G is a special case of SDG if we set $\sigma^i_{{\bf h}^{i}} = 0$. 
Thus, Theorem \ref{thm:1} is also applicable for SDG. 
It would be better if we can theoretically explain the improvement of SDG on G about UAP. 
Unfortunately, we cannot obtain such a kind of theorem at this time. 
As we known, the key spirits of investigating approximation ability are: the functional space to which the target belongs (denoted by $\Omega_t$), and the functional space which the approximation strategy can express (denoted by $\Omega_a$). 
For the practical problems, e.g., generating images, we have less knowledge of $\Omega_t$. 
We can only impose some basic assumptions on $\Omega_t$, like Assumption \ref{assump1}.1 and Assumption \ref{assump1}.2. 
These assumptions cannot characterize the complexity of the function in a real $\Omega_t$. 
For instance, generating images like ImageNet is much more difficult than generating images like MNIST.
If we cannot correctly define $\Omega_t$, it wound be hard or even impossible to theoretically compare the approximation ability of different $\Omega_a$ from different approximation strategies, i.e., the $\Omega_a({\rm G})$ by traditional G or the $\Omega_a({\rm SDG})$ by SDG. Apparently, we can assume some prior conditions on the complexity of $\Omega_t$ to obtain a theorem that satisfies our SDG, e.g., the density function in $\Omega_t$ has the form of production of Gaussian kernels. 
However, we do not think such kind of theorem is meaningful since the prior conditions are indeed unverifiable or overestimated. 
Thus, here we ignore the discussion on the theoretical investigation on SDG's improvement on G about UAP. 
However, we empirically compare the approximation ability of SDG and G, see the following experiments on a synthetic dataset. 
These results may conjecture that: given the fixed model architecture, $\Omega_a({\rm SDG})$ can approximate 
any function in $\Omega_a({\rm G})$, while $\Omega_a({\rm G})$ cannot approximate some functions in $\Omega_a({\rm SDG})$. 

\textbf{Reparameterization trick}. Unlike for the training of the weights in the traditional generator, stochastic data generation trains the hyperparameters of the conditional probability distributions. In the loss function, the expectation with respect to a distribution $\mathcal{N}(\mu^i_{{\bf h}^{i}}, \sigma^i_{{\bf h}^{i}})$ with parameters $\mu^i_{{\bf h}^{i}}, \sigma^i_{{\bf h}^{i}}$ has to be computed. However, it is difficult to calculate this expectation directly due to an indirect dependence on the parameters of distribution over which the expectation is taken; i.e., the traditional back-propagation gradients cannot flow directly~\cite{Rezende2014Stochastic}. Benefiting from the Gaussian reparameterization trick, our approach can be easily implemented; i.e., $p({\bf h}^{i+1} | {\bf h}^{i}) = \mu^i_{{\bf h}^{i}} + \sigma^i_{{\bf h}^{i}} \times \epsilon$, where $\mu^i_{{\bf h}^{i}}$ and $\sigma^i_{{\bf h}^{i}}$ depend on the previous layer and $\epsilon \sim \mathcal{N}(0, I)$. 
That means, in the feedforward process, we can sample ${\bf h}^{i+1}$ by instead sampling a normal Gaussian $\epsilon$. 
Then, in the back-propagation, we can easily obtain the gradients of loss function on $\mu^i_{{\bf h}^{i}}$ and $\sigma^i_{{\bf h}^{i}}$. 
We refer to~\cite{Kingma2014Auto,Rezende2014Stochastic} for the other details of reparameterization trick. 

\section{Experiments}
We evaluate the approximation ability of our proposed SDG and traditional G on a synthetic dataset and practical datasets. We interpret many of the results from the synthetic dataset in detail in Section \ref{subsec-synth} and find that these observations are also applicable to the practical datasets: MNIST, CIFAR10, and CELEBA in Section \ref{subsec-prac}.

\subsection{Synthetic dataset}\label{subsec-synth}
We seek to compare the approximation ability of G and SDG, that is, whether G can approximate the data distribution generated by SDG, or whether SDG can approximate the data distribution generated by G. 

\textbf{Dataset description}.
Now, we construct the synthetic dataset by an artificial G and SDG. 
For a traditional G with one hidden layer, we design a $D$-dimensional synthetic Data1 and Data2 as follows:
\begin{equation*}
X = A_{2}z_{2} + b_{2}, z_{2} = A_{1}z_{1} + b_{1}, z_1\sim \mathcal{N}(0,1.0*I_{d}).
\end{equation*}
where $A_1\sim \mathcal{N}(0, 1.0*I_{N_{1} \times d})$, $b_1\sim \mathcal{N}(0, 0.0001*I_{N_{1}})$, $A_2\sim \mathcal{N}(0, 1.0*I_{D \times {N_{1}}})$, $b_2\sim \mathcal{N}(0, 0.0001*I_{D})$, $d = 2$, $D = 10$ and $N_{1}$ is the number of neurons in the hidden layer of G. 
For Data1, $N_{1} = 100$, and for Data2, $N_{1} = 200$. 
It is observed from the above construction process that these two datasets are essentially Gaussian distributions. The Data1 and Data2 were designed to study the effect of an increase in the number of network parameters on the approximation ability for a traditional generator. 
For Data3, we use one hidden layer SDG with the same architecture with above G and $N_1=100$ to generate synthetic data:
\begin{equation*}
X = A_{2}z_{2} + b_{2}, z_{2} = \mu + \sigma\times\epsilon + b_{1}, 
\end{equation*}
where $\mu = A_{11}z_{1}$, $\sigma = A_{12}z_{1}$, $z_1\sim \mathcal{N}(0,1.0*I_{d})$, $\epsilon\sim \mathcal{N}(0,1.0*I_{N_{1}})$,
$A_{11}$ and $A_{12} \sim \mathcal{N}(0, 1.0*I_{{N_{1}} \times d})$, $b_1\sim \mathcal{N}(0, 0.0001*I_{N_1})$, $A_2\sim \mathcal{N}(0, 1.0*I_{D\times N_1})$, $b_2\sim \mathcal{N}(0, 0.0001*I_{D})$, $d = 2$, $D = 10$.
We randomly sample 10,000 points from the above distribution as a dataset. 

\textbf{Experimental setup}.
We use the artificial G and SDG of a constructing dataset for NSGAN~\cite{goodfellow2014generative}, respectively, named NSGAN and SDG-NSGAN. 
We use NSGAN-100 and NSGAN-200 to denote the NSGAN models with the number of single hidden layer neurons of $100$ and $200$, corresponding to Data1 and Data2, respectively. 
The above three models (NSGAN-100, NSGAN-200 and SDG-NSGAN) use the same discriminator architectures of a fully connected neural network with leaky ReLU activations: 10-100-1. 
We can check that NSGAN-200 has twice as many parameters as NSGAN-100. 
In SDG, a prior conditional probability is imposed between the input layer and the first fully connected layer and set to follow the Gaussian distribution; i.e., $p({\bf h}^{1}|{\bf z})\sim \mathcal{N}(\mu, \sigma)$, where the distribution hyperparameters $\mu$ and $\sigma$ depend on the input layer, similar to the design of Data3. 
Then, the total number of the parameters in SDG-NSGAN is between those of NSGAN-100 and NSGAN-200. 
The reparameterization trick is used to update the distribution hyperparameters. 
We set the prior distribution over the latent space to a Gaussian distribution $p_{\bf z}:=\mathcal{N}(0, I)$. 
We choose Adam as the optimization algorithm to update the parameters in these experiments, set $\beta_1 = 0.5$ and set the learning rate to $2 \times 10^{-4}$. 
We set the minibatch size to $64$ and the maximum number of generator iterations to $50$K. 

\textbf{Experimental results}.
Table \ref{tab:js} shows the results for the Jensen-Shanno (JS) divergence between the data distribution of the trained generator and the ground truth data distribution. 
Since Data1 and Data2 are essentially Gaussian and the data distribution is simple, the JS divergences obtained by the three models are very small (as shown in the first two rows of Table \ref{tab:js}), indicating that SDG can approximate the data distribution generated by G with competitive accuracy of G itself. 
For Data3, we can obtain better accuracy by increasing the model size of G, as observed from the results for NSGAN-100 and NSGAN-200. 
However, these two results are still not comparable with that of SDG, which means that G cannot approximate the data distribution generated by SDG. 
Since the total number of parameters in these three models is in the order of NSGAN-100 $<$ SDG-NSGAN $<$ NSGAN-200, we can further conclude that the introduced SDG can highly enhance the approximation ability, rather than increasing it merely by increasing the model size.

\begin{figure*}[htbp]
	%	\centering
	\begin{center}
		\includegraphics[scale=0.6]{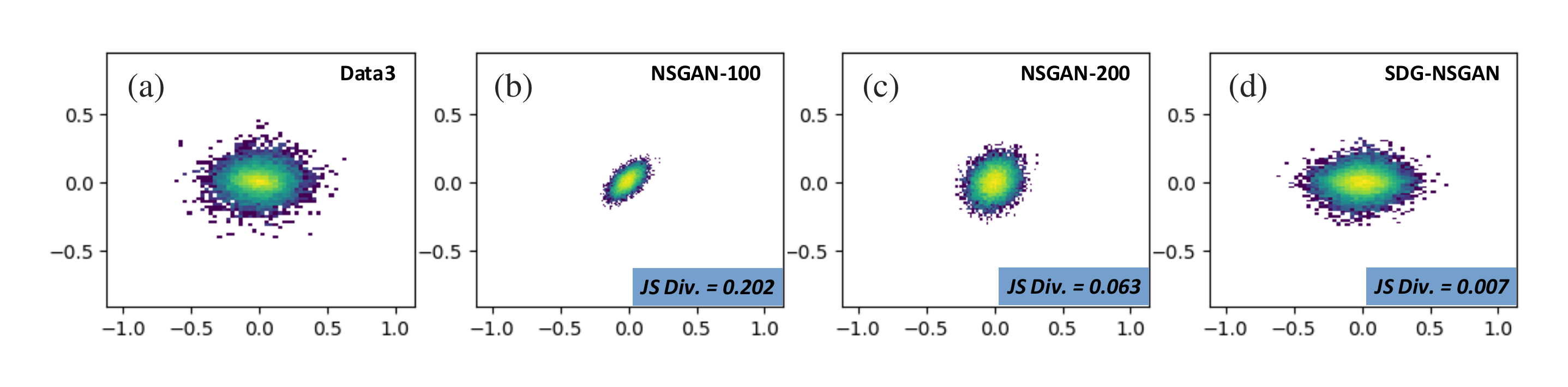}
		\vspace{-0.8cm}
	\end{center}
	\caption{2D PCA visualization sampling results of the data distributions: (a) Data3, (b) NSGAN-100, (c) NSGAN-200 and (d) SDG-NSGAN, where JS Div. is Jensen-Shanno divergence distance between the data distribution of the generative model and the original data distribution.} \label{fig_pca}
	%\vspace{-0.2cm}
\end{figure*}

\begin{table}
	\caption{Jensen-Shanno divergence (JS Div.) results on the synthetic dataset. We report the mean value and standard deviation under $20$-time data generation with different seeds. These results clearly demonstrate that: SDG-NSGAN can approximate Data1 and Data2 with competitive or even better behavior than NSGAN-100/NSGAN-200, while NSGAN-100/NSGAN-200 cannot approximate Data3. We remark that Data1, Data2, and Data3 are constructed using the generator in NSGAN-100, NSGAN-200, and SDG-NSGAN, respectively.}
	\label{tab:js}
	\centering
	\renewcommand\tabcolsep{3.5pt}
	\begin{tabular}{lrrr}  
		\toprule
		JS Div.		& NSGAN-100		& NSGAN-200 		& SDG-NSGAN\\
		\midrule
		Data1    & 0.0061$\pm$0.0018 & 0.0111$\pm$0.0063	& {\bf 0.0051$\pm$0.0013}    \\
		Data2    & 0.0049$\pm$0.0019 & 0.0066$\pm$0.0019 & {\bf 0.0067$\pm$0.0054}    \\
		Data3   	& {\bf 0.2072$\pm$0.1581} & {\bf 0.1925$\pm$0.1803} & 0.0163$\pm$0.0063    \\
		\bottomrule
	\end{tabular}
	%\vspace{0.2cm}
\end{table}
Figure \ref{fig_pca} shows the two-dimensional principal component analysis (PCA) visualization results of the three models on Data3. It is also observed from Figure \ref{fig_pca} that the approximation ability of SDG for the data distribution is improved and that the details of the original distribution can be captured. 

\subsection{Practical dataset}\label{subsec-prac}
Currently, generative adversarial networks are widely used in the field of image generation. 
The SDG approach can be widely used in a variety of original or improved GANs, like \cite{Brock2019Large,Karras2018A,Karras2019,Zhang2018}. 
Considering that the main objective of this paper is to show that SDG has better approximation ability than traditional G, i.e., we only consider the generator of GANs, not to directly generate images of better quality, we compare SDG and G on four GANs: NSGAN, DRAGAN, WGANGP and LSGAN for three image datasets: MNIST, CIFAR10, and CELEBA. 
Moreover, to further illustrate the difference between our SDG and self-modulation method \cite{Chen2019On} that directly impose the noise on the hidden layers as a function of input noise ${\bf z}$, we introduce self-modulation to NSGAN and DRAGAN, named SM-NSGAN and SM-DRAGAN, respectively. 

\textbf{Experimental setup}. To ensure a fair comparison, we use the same generator and discriminator architecture as those in \cite{Lucic2018Are}. 
In particular, in the SDG model, a prior conditional probability assumption is introduced between the input layer and the first fully connected layer with $512$ neurons. It is easy to check that the difference in parameters between our SDG method and other G of above four GANs comes from the first two layers.
Note that the number of neurons in the first two fully connected layers in \cite{Lucic2018Are} are $1024$ and $128\times h/4 \times w/4$, respectively, 
where $h$ and $w$ are the quantities related to the dataset, specifically, $h\times w$ is $28\times28$, $32\times32$, and $64\times64$ for MNIST, CIFAR10 and CELEBA, respectively.
We fix the latent code size to $64$ and the prior distribution over the latent space to be Gaussian $\mathcal{N}(0,I)$. 
It is easy to show that our SDG model will reduce $512 \times 128\times h/4 \times w/4$ weights, which is $3211264$, $4194304‬$, or $16777216$ for MNIST, CIFAR10 or CELEBA, respectively. 
We choose Adam as the optimization algorithm to update the parameters in these experiments. We set the batch size to $64$ and perform optimization for $20$ epochs on MNIST, for $40$ epochs on CELEBA and for $100$ epochs on CIFAR10. 
We refer the reader to \cite{Lucic2018Are} for the other details of the experimental setup. 
To control the number of parameters, we only apply self-modulation to the first batch normalization layer with input-dependent parameters $\beta({\bf z})$ and $\gamma({\bf z})$. As mentioned in \cite{Chen2019On}, a same one-hidden layer feed-forward network with ReLU activation is applied to the generator input ${\bf z}$: 64-64-1024, and accordingly will introduce additional $(64 \times 64 + 64 + 64 \times 1024) \times 2= 139392$ weights.  

\textbf{Experimental results}. We now discuss the quantitative measurement results using Fr$\acute{e}$chet Inception Distance (FID) \cite{Heusel2017} and Inception Score (IS) \cite{Salimans2016Improved}. 
FID is computed by considering the difference in embedding of true and fake data. 
We report FID and IS using the same strategy as that used in~\cite{Lucic2018Are}, with two stages: we first run a large-scale search on $100$ sets of hyperparameters, and select the best model. 
Then, we rerun the training process of the selected best model $10$ times with different initialization seeds to report the mean value and standard deviation of FID and IS. As shown in Table \ref{tab:fid}, NSGAN, DRAGAN, WGANGP and LSGAN are the baseline results from \cite{Lucic2018Are}. 
The corresponding GANs with SDG, denoted by SDG-NSGAN, SDG-DRAGAN, SDG-WGANGP and SDG-LSGAN can outperform the traditional models except for the MNIST dataset and LSGAN for the CELEBA dataset. 
This may be because of the simple distribution of MNIST due to which NSGAN performs very well, leading to a limited improvement by the introduction of SDG. 
%We note that our SDG leads to little change in the model architectures. 
It is worth noting that our SDG requires a smaller network architecture and SM-NSGAN or SM-DRAGAN will introduce a larger network architecture to achieve similar results.

\begin{table}
	\caption{Best FID achieved by each model on three datasets.} 
	\label{tab:fid}
	\centering
	\begin{tabular}{lrrr}  
		\toprule
		FID			& MNIST		&CIFAR10	&CELEBA\\
		\midrule
		NSGAN			&6.8$\pm$0.5	&58.5$\pm$1.9	&55.0$\pm$3.3\\
		SM-NSGAN			&\textbf{5.9$\pm$0.5}	&56.0$\pm$2.4	&54.5$\pm$3.2\\
		\textbf{SDG-NSGAN}		&7.0$\pm$0.5	&\textbf{54.3$\pm$1.9}	&\textbf{54.5$\pm$1.6}\\
		\midrule
		DRAGAN		&\textbf{7.6$\pm$0.4}	&69.8$\pm$2.0	&42.3$\pm$3.0\\
		SM-DRAGAN		&10.0$\pm$1.7	&\textbf{55.9$\pm$1.8}	&35.8$\pm$1.2\\
		\textbf{SDG-DRAGAN}	&9.2$\pm$1.1	&56.5$\pm$1.1	&\textbf{34.4$\pm$2.1}\\
		\midrule
		%WGAN		&\textbf{6.7$\pm$0.4}	&\textbf{55.2$\pm$2.3}	&\textbf{41.3$\pm$2.0}\\
		%\textbf{SDG-WGAN}	&7.93$\pm$0.91	&56.59$\pm$2.03	&42.21$\pm$3.91\\
		WGANGP		&20.3$\pm$5.0	&55.8$\pm$0.9	&30.0$\pm$1.0\\
		\textbf{SDG-WGANGP}	&\textbf{11.7$\pm$0.6}	&\textbf{51.5$\pm$2.4}	&\textbf{27.9$\pm$1.5}\\
		\midrule
		LSGAN		&\textbf{7.8$\pm$0.6}	&87.1$\pm$47.5	&\textbf{53.9$\pm$2.8}\\
		\textbf{SDG-LSGAN}	&8.6$\pm$1.1	&\textbf{60.6$\pm$2.0}	&59.6$\pm$4.6\\
		\bottomrule
	\end{tabular}
	%\vspace{0.2cm}
\end{table}

\begin{table}
	\caption{Best IS achieved by each model on three datasets.} 
	\label{tab:is}
	\centering
	\begin{tabular}{lrrr}  
		\toprule
		IS			& MNIST		&CIFAR10	&CELEBA\\
		\midrule
		\text{NSGAN}		&2.23$\pm$0.02	&6.10$\pm$0.18	&1.85$\pm$0.02\\
		\textbf{SDG-NSGAN}		&\textbf{2.23$\pm$0.03}	&\textbf{6.28$\pm$0.17}	&\textbf{1.87$\pm$0.02}\\
		\midrule
		\text{WGANGP}	&2.27$\pm$0.02	&5.45$\pm$0.37	&2.22$\pm$0.05\\
		\textbf{SDG-WGANGP}	&\textbf{2.28$\pm$0.01}	&\textbf{6.24$\pm$0.13}	&\textbf{2.39$\pm$0.04}\\
		\midrule
		\text{DRAGAN}		&2.26$\pm$0.03	&5.71$\pm$0.06	&2.06$\pm$0.02\\
		\textbf{SDG-DRAGAN}		&\textbf{2.30$\pm$0.03}	&\textbf{5.85$\pm$0.07}	&\textbf{2.11$\pm$0.04}\\
		\bottomrule
	\end{tabular}
\end{table}
Table \ref{tab:is} shows the comparison results of IS on three models: NSGAN, WGANGP and DRAGAN. 
The corresponding baseline results of IS are reproduced by using the same experimental settings as in \cite{Lucic2018Are}. As shown in Table \ref{tab:is}, for IS measurement, the performance of SDG method is better than other corresponding GANs.

\begin{figure*}[htbp]
	%	\centering
	\begin{center}
		\includegraphics[scale=0.80]{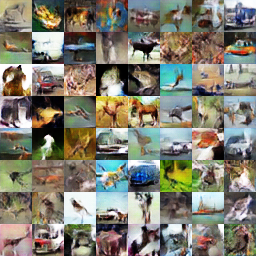}~~~~~
		\includegraphics[scale=0.40]{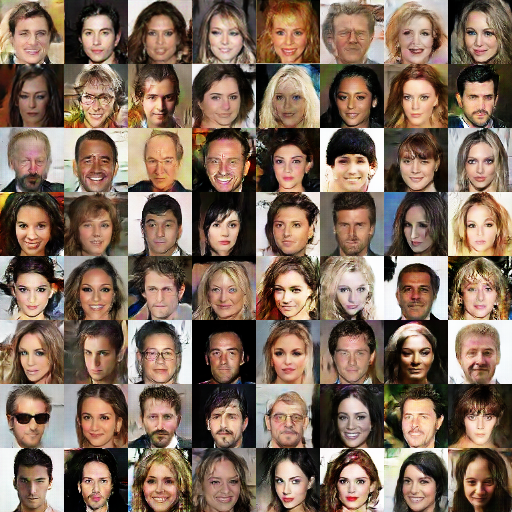}
		%\vspace{-0.2cm}
	\end{center}
	\caption{Generative images produced by our SDG generator for the CIFAR10 dataset (left) at 32$\times$32 and CELEBA dataset (right) at 64$\times$64.}\label{fig_cifar10}
\end{figure*}

Finally, we provide some qualitative results on CIFAR10 and on CELEBA in Figure \ref{fig_cifar10}, respectively. We find that for CIFAR10, due to the influence of the resolution of the dataset itself, the generated images are relatively blurry, but it can still be observed that the generated images have rich diversity. As shown in Figure \ref{fig_cifar10} (right), the generated images for CELEBA have high quality and diversity. 
\section{Conclusions}
This paper focuses on GANs' approximation property.
We first theoretically prove that the generator of GANs can universally approximate the potential data distribution. 
Then, we propose an approach named stochastic data generation (SDG) to enhance GANs' approximation ability by introducing a prior Gaussian distribution on the conditional probability between the layers of GANs generator. 
By using the reparameterization trick, we can easily update the hyperparameters of the conditional probability distribution based on back-propagation. 
The experimental results on the synthetic dataset and the practical dataset verify the SDG improvement. 

In the future, as remarked above, the weakened form of Assumption \ref{assump1}.3 will undergo further theoretical and empirical investigations.  Furthermore, the general theoretical proof of UAP for GANs generator will be studied from following aspects: modifying the construction strategy  for general data distribution, directly estimating the error between the joint distribution and the product of marginal distributions, or proving it from the perspective of functional analysis. In addition, the property of the functional space to which the target belongs is an open aspect and requires more theoretical research. 

\section*{Acknowledgements}
This work was supported in part by the Innovation Foundation of Qian Xuesen Laboratory of Space Technology, and in part by Beijing Nova Program of Science and Technology under Grant Z191100001119129. 

% ---- Bibliography ----
\bibliographystyle{ieee_fullname}
\bibliography{reference-File}

\clearpage
{\bf Appendix: the proof of Lemma~\ref{lem:2}}.
\begin{proof}
	We first consider the 1D case, i.e., $d = D = 1$. We will provide a constructive proof. 
	The key idea comes from the approximation for a discrete distribution. 
	Suppose the discrete data distribution $p_{X}(x_i) = y_i$, $i=1,\cdots,m$, and $\sum_i y_i = 1$. 
	Letting $y_0=0$, we can then define  a piecewise constant function $Q(z)$, $z\in(0,1)$ as $i=1,\cdots,m$, 
	\bea\label{discrete_cons}
	Q(z) = x_i, \quad \sum_{j=0}^{i-1}y_j < z < \sum_{j=0}^{i}y_j.
	\eea
	It is easy to show that 
	\bena
	P_{Q(z)}(x) &= \mathbb{E}_{{\bf z} \sim U(0,1)}\,\delta (x - Q(z))\\ 
	&=\sum_{i=1}^{m}\delta(x-x_i)y_i =p_{X}(x_i).
	\eena
	This means that we can construct a function $Q(z)$ with the output that can produce a target discrete data distribution. 
	
	Following a similar approach for \eqref{discrete_cons}, we can construct function $Q(z)$ for a continuous data distribution. By Assumption \ref{assump1}.1, suppose the compact set $K=[k_1,k_2]$. 	
	Now, we define
	\bea\label{continuous_cons}
	z(\lambda) &= \int_{k_1}^{k_1+\lambda(k_2-k_1)} p_X(x) dx, \\
	Q(\lambda) &= k_1+\lambda(k_2-k_1),
	\eea
	where $\lambda\in (0,1)$. Since $\int_{k_1}^{k_2} p_X(x) dx = 1$, we have $z(\lambda) \in (0,1)$. 
	It is easy to show that 
	\bena
	p_{Q(z)}(x) &= \mathbb{E}_{{\bf z} \sim U(0,1)}\,\delta (x - Q(z)) = \int_{0}^{1}\delta(x-Q(z))dz \\
	&= \int_{0}^{1}\delta(x-k_1-\lambda(k_2-k_1)){{(k_2-k_1)}}\\ 
	&\quad \quad p_X(k_1+\lambda(k_2-k_1)) d\lambda \\
	&=p_X(x).
	\eena
	
	By the definition in \eqref{continuous_cons}, we have $\forall\, \lambda_1 < \lambda_2$; 
	let $z_1 = z(\lambda_1), z_2 = z(\lambda_2), Q_1 = Q(\lambda_1), Q_2 = Q(\lambda_2)$, and then, 
	\bena
	z_1 - z_2 = \int_{k_1+\lambda_1(k_2-k_1)}^{k_1+\lambda_2(k_2-k_1)} p_X(x) dx,
	\eena
	considering that $p_X(x)$ is a continuous function, we obtain
	\bena
	z_2 - z_1 = (\lambda_2-\lambda_1) p_X(\hat x),
	\eena
	where $\hat x$ is an implicit variable located in $[k_1+\lambda_1(k_2-k_1),k_1+\lambda_2(k_2-k_1)]$ 
	and $p_X(\hat x) \neq 0$. 
	In addition, we also have $Q_2 - Q_1=(\lambda_2-\lambda_1)(k_2-k_1)$, such that 
	\bena
	|Q_2-Q_1| = |z_2 - z_1| \cdot |k_2-k_1|/p_X(\hat x) , 
	\eena
	which means that $Q(z)$ is a continuous function. 
	
	Actually, this strategy can be easily extended to the proof for a general prior distribution on $z$ for the 1D case. Suppose $z \sim p_{Z}(z)$ with $z \in [z_{min}, z_{max}]$. We define the probability distribution function of $x$ and $z$ as 
	\bena
	F_{X}(x) = \int_{k_1}^{x}\,p_{X}(t)dt,\quad
	F_{Z}(z) = \int_{z_{min}}^{z}\,p_{Z}(t)dt,
	\eena
	and by Assumption \ref{assump1}, we have $p_{X}(t) > 0$, indicating that $F_{X}$ is inversable and that $F_{X}^{-1}$ is continuous. Then, we can define $Q(z) = F_{X}^{-1}(F_{Z}(z)): [z_{min}, z_{max}] \rightarrow [k_{1}, k_{2}]$, which is continuous and 
	\bena
	p_{Q(z)}(x) &= \mathbb{E}_{z \sim p_{Z}(z)}\,\delta (x - Q(z)) \\
	&= \int_{z_{min}}^{z_{max}}\delta(x-Q(z))p_{Z}(z)dz \\
	&= \int_{0}^{1}\,\delta (x - Q(z))dF_{Z}(z)\\
	&= \int_{k_{1}}^{k_{2}}\,\delta (x - Q(z))p_X(Q(z)) d Q(z)\\
	&=p_X(x).
	\eena
	
	We then consider the high-dimensional case, $d > 1$. We will take the 2D case of $d = 2$ as an example because similar results will be obtained for higher dimensions. By Assumption \ref{assump1}, suppose that $p_{\bf X}({\bf x}) = p_{\bf X}(x^1, x^2)$. We define ${\bf z} = (z^1, z^2)^{T}$. Since $K$ is a compact set, we assume the corresponding compact sets of potential variables $x^1,x^2$ are $[k_1^1,k_2^1],[k_1^2,k_2^2]$. 	Now, we can define 	
	\begin{gather}\nonumber
	\begin{pmatrix}
	z^1(\lambda^1)\\z^2(\lambda^2)
	\end{pmatrix}
	=
	\begin{pmatrix}
	\int_{k_1^1}^{k_1^1+\lambda^1(k_2^1-k_1^1)} p_{X_1}(x^1) dx^1\\\int_{k_1^2}^{k_1^2+\lambda^2(k_2^2-k_1^2)} p_{X_2}(x^2) dx^2
	\end{pmatrix}
	,\\
	%\end{gather*}
	%\begin{gather*}
	\begin{pmatrix}\nonumber
	Q^1(\lambda^1)\\Q^2(\lambda^2)
	\end{pmatrix}
	=
	\begin{pmatrix}
	k_1^1+\lambda^1(k_2^1-k_1^1)\\k_1^2+\lambda^2(k_2^2-k_1^2)
	\end{pmatrix}
	,
	\end{gather}
	%	\bea
	%	(z(\lambda^1), z(\lambda^2)) &= (\int_{k_1^1}^{k_1^1+\lambda^1(k_2^1-k_1^1)} p_1(x_1) dx_1, \int_{k_1^2}^{k_1^2+\lambda^2(k_2^2-k_1^2)} p_2(x_2) dx_2), \\
	%	(Q(\lambda^1), Q(\lambda^2))&= (k_1^1+\lambda^1(k_2^1-k_1^1), k_1^2+\lambda^2(k_2^2-k_1^2)),
	%	\eea
	where $\lambda^1, \lambda^2\in (0,1)$, $p_{X_1}(x^1) = \int_{k_1^2}^{k_2^2}p_{\bf X}(x^1, x^2) dx^2$ and $p_{X_2}(x^2) = \int_{k_1^1}^{k_2^1}p_{\bf X}(x^1, x^2) dx^1$ are the marginal probability densities. 
	Define $Q({\bf z}) = (Q^1(\lambda^1),Q^2(\lambda^2))^T$. 
	It is also easy to show that 
	\bena
	&p_{Q(\bf z)}({\bf x}) = \mathbb{E}_{{\bf z} \sim U^2(0,1)}\,\delta (\mathbf{x} - Q({\bf z})) \\
	&= \int_{0}^{1}\int_{0}^{1}\delta({\bf x} - Q({\bf z}))dz^1dz^2 \\
	%&= \int_{0}^{1}\int_{0}^{1}\delta(x^1-Q(\lambda^1), x^2 - Q(\lambda^2))\\
	%&\quad \quad (k_2^1-k_1^1)p_{X_1}(k_1^1+\lambda^1(k_2^1-k_1^1))\\
	%&\quad \quad (k_2^2-k_1^2)p_{X_2}(k_1^2+\lambda^2(k_2^2-k_1^2))d\lambda^1d\lambda^2 \\
	&= \int_{0}^{1}\delta(x^1-Q(\lambda^1)) (k_2^1-k_1^1)p_{X_1}(k_1^1+\lambda^1(k_2^1-k_1^1))d\lambda^1\\
	&\quad\int_{0}^{1}\delta(x^2 - Q(\lambda^2))(k_2^2-k_1^2)p_{X_2}(k_1^2+\lambda^2(k_2^2-k_1^2))d\lambda^2 \\
	&= p_{X_1}(x^1)p_{X_2}(x^2)\\
	&=p_{\bf X}({\bf x}).
	\eena
	where the last equality holds based on Assumption \ref{lem:1}.3.
	Since we assume that the random potential variables $x^1,x^2$ are independent of each other, 
	we have $p_{\bf X}(x^1,x^2) = p_{X_1}(x^1)p_{X_2}(x^2)$. 
	Furthermore, similar to the 1D case, we can also prove that $Q({\bf z})$ is a continuous function. 
	
	Now, we finish the proof. 
\end{proof}
\end{document}